\documentclass[10pt,journal,compsoc]{IEEEtran}
\usepackage{amsmath,amsfonts}
\usepackage{algorithmic}
\usepackage{algorithm}
\usepackage{array}
\usepackage[caption=false,font=normalsize,labelfont=sf,textfont=sf]{subfig}
\usepackage{textcomp}
\usepackage{stfloats}
\usepackage[hyphens]{url}
\usepackage{verbatim}
\usepackage{graphicx}
\usepackage[nocompress]{cite}
\usepackage{amssymb}
\usepackage{multirow}
\usepackage{xcolor}
\usepackage{booktabs}
\usepackage[colorlinks=true,linkcolor=blue,citecolor=blue,urlcolor=blue]{hyperref}
\usepackage{orcidlink}

\begin{document}

\title{SLA-Aware Task Offloading in the IoT Edge-Cloud Continuum:\\ Class-Weighted Deep Reinforcement Learning\\ Against an Exact Optimum}

\author{Shilpa~Golla\,\orcidlink{0009-0005-3127-9555}, Tharun~Kumar~Korine~Palli\,\orcidlink{0009-0000-2474-3691}, and Urmila~Bandaru\,\orcidlink{0000-0001-9712-4242}
\IEEEcompsocitemizethanks{\IEEEcompsocthanksitem This work has been submitted to the IEEE for possible publication. Copyright may be transferred without notice, after which this version may no longer be accessible.
\IEEEcompsocthanksitem S.~Golla is with the Deggendorf Institute of Technology (DIT), Cham, Germany.\protect\\ E-mail: shilpa.golla@stud.th-deg.de
\IEEEcompsocthanksitem T.~K.~Korine~Palli is with the Technologietransferzentrum Kitzingen (TTZ-KT), Technische Hochschule W\"urzburg-Schweinfurt (THWS), Kitzingen, Germany.\protect\\ E-mail: tharunkumar.korinepalli@thws.de
\IEEEcompsocthanksitem U.~Bandaru is an independent researcher.\protect\\ E-mail: urmila913@gmail.com
\IEEEcompsocthanksitem This work received no external funding. Shilpa Golla and Tharun Kumar Korine Palli contributed equally to this work. The work was supervised by Tharun Kumar Korine Palli and Urmila Bandaru. \emph{(Corresponding author: Shilpa Golla.)}}}

\IEEEtitleabstractindextext{
\begin{abstract}
Deep reinforcement learning (RL) is often used to decide where Internet of Things (IoT) tasks with different latency limits run in the edge-cloud continuum, yet some methods give every missed task the same penalty and few compare their agents with an optimal policy. This benchmark study builds four offloading systems small enough to solve exactly by dynamic programming and measures reward and learner against the optimum. A bonus for a met limit makes each miss cost its penalty plus the bonus, which shifts the class weights of published per-class rewards. A class-blind penalty stays 65\% to 118\% above the optimal cost even with a perfect learner, while a corrected per-class penalty comes within 0.13\% of the optimum. With a per-class reward, tuned proximal policy optimization (PPO) agents come within 9\% to 11\% of the optimum in the smallest system, and the corrected one beats a tuned reservation rule in 9 of 10 seeds. In larger systems the agents stay 41\% to 76\% above the optimum, and a tuned rule beats their mean. Training ten times longer does not close this gap, and a PPO-Lagrangian agent with reachable targets does no better than the per-class reward.
\end{abstract}

}

\maketitle
\IEEEdisplaynontitleabstractindextext
\IEEEpeerreviewmaketitle

\IEEEraisesectionheading{\section{Introduction}\label{sec:intro}}
\IEEEPARstart{I}{oT} devices in factories, buildings and hospitals create a steady stream of computing tasks, and each task can run on a small edge server near the device or on a large cloud server far away \cite{shi2016,satyanarayanan2017}. This range of places is called the edge-cloud continuum. Choosing where each task runs is called task offloading \cite{mao2017,mach2017}. A service level agreement (SLA) sets a latency limit for each type of task, and a task that finishes late violates it. IoT tasks are not equally urgent: latency-critical applications suit edge servers, which answer within tens of milliseconds, while latency-tolerant ones can use the cloud at more than 100\,ms \cite{mao2017}. Both compete for the same edge servers, so when the edge is busy the offloading policy, the rule that picks a place for each task, must decide which tasks to protect. For the operator of an edge-cloud platform this is a resource management problem. The edge capacity is shared, and the policy decides which SLA class gets it when demand reaches capacity.

Deep reinforcement learning (RL) \cite{mnih2015} is a common way to learn this policy \cite{hortelano2023}: the agent looks at the system, places a task and gets a reward. Two gaps remain. First, some RL offloading methods give every missed task the same penalty, whatever its class \cite{nieto2024}, so the average can look good while critical tasks keep missing their limit. A recent method gives each class its own penalty \cite{yamsani2026}, but it also changes the state, the action set and the network, and no ablation run changes only the class penalties, so the effect of the penalty alone is unknown. Second, few offloading studies compare their agent with an optimum. Some compare with the best decision for a single time frame \cite{huang2020} or a single task graph \cite{wang2021}, found by exhaustive search, and exact Markov decision process solutions exist for some small queued offloading problems \cite{mao2017}. However, none of the deep RL offloading studies cited here compares its agent with the optimal policy of a system with queues. Most compare only with simple rules and other agents, so it is not known how much better the policy could be.

This paper builds offloading systems that are small enough to solve exactly but keep the hard parts of the problem: up to four edge nodes of different speeds, up to three task classes, random background traffic and a cloud that meets some limits only by chance. For such systems, dynamic programming (DP) \cite{puterman1994} gives the optimal policy, the Pareto frontier of miss rates, and the exact expected score of any policy that is a table over the states, including a trained network. It also gives the policy that a perfect learner of a given reward would reach, called the DP policy of that reward, which separates what a reward asks for from how well an agent learns it. The agents are trained with proximal policy optimization (PPO), deep Q-network (DQN), advantage actor-critic (A2C) and a Lagrangian variant of PPO.

The paper is a benchmark and methodology study that proposes no new learner but measures exactly how far common reward designs and standard learners are from the optimum. It makes four contributions. First, it gives an exact model of a family of multi-class edge-cloud offloading systems, checked against the simulator. Second, it shows that a reward with a bonus for a met limit ranks misses by the penalty plus the bonus, an effect that is short to prove but not visible in published designs. It explains why, for a perfect learner, a class-blind penalty minimizes the overall miss rate instead of the SLA cost, and it gives a simple rule to correct the penalties. Third, it reports a study with equal tuning budgets and 10 seeds for PPO, DQN and A2C in the nominal system and for PPO in three larger systems, with one run ten times longer than the main budget. The agents are compared with rules tuned on the exact model, with rewards based on two published designs, with a constrained PPO-Lagrangian agent and with the optimum, and PPO is tested under five changes of the nominal system. Fourth, the code, the results, the policy tables and the trained PPO networks are released.

\section{Related Work}
\label{sec:related}

\emph{Deep RL for task offloading.} Surveys treat offloading as a choice of where a task runs, made together with the radio and computing resources it uses \cite{mao2017,mach2017}. Many studies learn this choice with deep RL, and a recent survey sorts them by use case, objective and RL algorithm \cite{hortelano2023}. They learn binary offloading decisions in wireless powered networks \cite{huang2020,chen2025}, offloading with energy and channel states \cite{chen2019,min2019}, fast adaptation by meta RL \cite{wang2021}, task scheduling with transmit power and computing resources set by optimization \cite{hu2021}, or offloading aided by reconfigurable intelligent surfaces \cite{zhang2024}. Tang and Wong \cite{tang2022} penalize dropped tasks with deadlines, Ullah et al. \cite{ullah2023} use the deadline slack of each task in the reward, and Binh et al. \cite{binh2024} penalize the time by which a task passes its deadline. Some of these works give each task or device its own deadline, but none has task classes whose misses cost different amounts. They treat all misses alike or fold all goals into one weighted score. This paper asks instead which class is protected when the edge is full.

\emph{Learning-based scheduling in cloud-edge systems.} Cloud-edge platforms also use deep RL to place and schedule tasks. Shuai et al. \cite{shuai2023} combine domain adaptation with deep RL to offload tasks from heterogeneous devices to distributed edge clouds, and test it with real devices. Chen et al. \cite{chen2024adprl} decide offloading and bandwidth together for dependent tasks with a dependency-aware reward, and Zou et al. \cite{zou2025} schedule dependent tasks with PPO while executors are created early and reused to cut startup delays. Xu et al. \cite{xu2025} use causal RL so that one offloading policy works on unseen task distributions without fine-tuning. Pandey and Kwon \cite{pandey2026} place serverless functions with a double dueling deep Q-network, while a scheduler inside each node protects deadline-sensitive requests. These studies aim at completion time, energy, dropped tasks, generalization or the share of completed tasks, and they compare their methods mainly with heuristic or learned baselines. Two of them also use an optimum. Chen et al. compare their agent, in one small example, with an offline schedule that knows all arrivals. Xu et al. compare with an exhaustive search over the offloading plan of a single task graph. Their agent has 11.8\% higher latency than this search, which becomes intractable for larger graphs. None of them computes the optimal policy of a system with random arrivals.

\emph{QoS- and SLA-aware offloading.} AlQerm and Pan \cite{alqerm2021} give each application a fixed priority and maximize one quality-of-experience score over all applications. Heidari et al. \cite{heidari2023} maximize one efficiency score that mixes delay, energy and failures, so both fold all tasks into one number. Kayal and Leon-Garcia \cite{kayal2024} split data-intensive task graphs and schedule the parts across edge, near-edge and cloud tiers to meet QoS and SLA needs at low latency or cost with a low-complexity myopic heuristic. Nieto et al. \cite{nieto2024} train actor-critic agents with three task classes in the state, but every class gets the same fixed penalty when a task fails. SLA-ORECS \cite{lan2024} uses PPO to split resources among SLA channels and linear programming to dispatch requests, not to place each task. Yamsani and Chenna Reddy \cite{yamsani2026} train a DQN-based scheduler with an SLA risk score in the state, Gold, Silver and Bronze penalty tiers that grow when a tier misses its target, and pruning of unsafe actions. They compare no policy with an optimum. This paper rebuilds their adaptive tiers, risk features and pruning as closely as their description allows. Georgiades et al. \cite{georgiades2026} offload vehicular tasks of three application classes with deadlines from 0.2 to 1\,s. Their PPO reward penalizes a missed deadline with the same weight for every class and adds Lagrangian pressure when the overall failure rate exceeds its target. Mohajer et al. \cite{mohajer2026} train multi-agent TD3 for SLA-aware offloading and slicing. Their shaped reward (Eq.~26) gives every task the same bonus for meeting one common delay limit, while their revenue and their Eq.~18 reward weight tasks by a priority from 1 to 5. Neither computes the optimal policy of a system with queues, and Georgiades et al. define regret only against a per-round oracle.

\emph{Reward design, constrained and multi-objective RL.} Ng et al. \cite{ng1999} show that only potential-based shaping terms added to the reward keep the optimal policy in every Markov decision process. A bonus for a met limit is not such a term, and Section~\ref{sec:rewardproof} shows how it moves the optimal policy when penalties differ by class. A reward that differs from the true goal can be gamed by the agent \cite{skalse2022}, and more capable agents often game it more \cite{pan2022}. Constrained Markov decision processes \cite{altman1999} treat such goals as limits. Constrained policy optimization \cite{achiam2017} tries to keep each policy update inside the limits, while Lagrangian methods \cite{ray2019,tessler2019} raise a penalty while a limit is violated. The adaptive tiers of \cite{yamsani2026} resemble these Lagrangian methods with a constant step size, and Section~\ref{sec:lagrangian} tests a PPO-Lagrangian agent whose targets can be met. Multi-objective RL can learn policies that cover many weightings \cite{roijers2013,hayes2022,yang2019}. In edge-cloud orchestration, Di Cicco et al. \cite{dicicco2025} train PPO over a grid of preference weights to approximate the Pareto front of acceptance, latency and energy, and compare it with heuristics. For the systems here, DP gives this front exactly. This paper does not propose a new learner. It shows exactly what fixed and adaptive penalties optimize and how far each lands from the optimum.

\emph{Optimal benchmarks and evaluation of deep RL.} For one pool of equal servers with no waiting room, several classes and the same service rate for every class, the optimal admission policy is a threshold rule that keeps capacity free for the more valuable classes \cite{miller1969}. With servers of different speeds, random background traffic and a cloud that sometimes meets a limit, no proven simple optimal form is known to the authors, so this paper computes the optimum numerically. For small Markov decision processes, DP gives the optimal policy and the exact value of any policy \cite{puterman1994}, while large ones usually need approximate DP \cite{powell2011adp}. Queueing models also guide allocation in cloud platforms. Okuyama and Inoue \cite{okuyama2026} derive closed-form latency approximations for GPU inference servers with dynamic batching, and the resulting server allocation comes close to the optimal allocation found by exhaustive simulation. PPO can come close to the optimum in queueing networks where the optimum is known \cite{dai2022queueing}. Deep RL results also change with the seed and with the evaluation protocol \cite{henderson2018,agarwal2021}.

\section{System Model}
\label{sec:model}

\subsection{System}
\begin{figure}[t]
\centering
\includegraphics[width=\columnwidth]{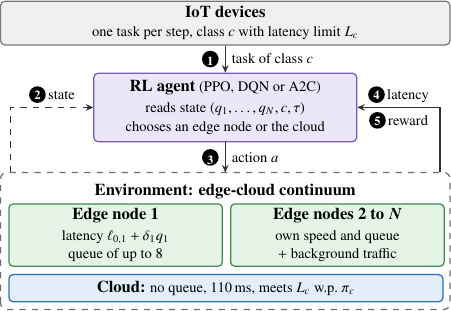}
\caption{Offloading loop of the system. Steps (1) to (5) are described in Section~\ref{sec:steps}. Each edge node has its own speed and a queue of up to 8 tasks. Background tasks from other devices join the shortest edge queue, which can be any node. The cloud has no queue and meets the limit of class $c$ with probability $\pi_c$.}
\label{fig:architecture}
\end{figure}

Fig.~\ref{fig:architecture} shows the system. Time runs in steps $t = 0, 1, \dots, T-1$, where $T$ is the number of steps in an episode. In every step, IoT devices send one task. The task belongs to class $c \in \{1,\dots,K\}$ with probability $p_c$, independently of all other tasks, where $K$ is the number of classes. A task of class $c$ must finish within its latency limit of $L_c$ milliseconds. Each task is placed on one of $N$ edge nodes or on the cloud.

Edge node $i$ holds a queue of $q_i$ waiting tasks. A task placed on edge node $i$ gets the latency
\begin{equation}
\ell = \ell_{0,i} + \delta_i\, q_i ,
\label{eq:latency}
\end{equation}
where $\ell_{0,i}$ is the processing time of the node and $\delta_i$ is the extra wait for each task already in its queue. The latency is known when the task is placed, so an edge task meets its limit exactly when $\ell \le L_c$. A queue holds at most 8 tasks. A task placed on a full queue gets the latency of Eq.~\eqref{eq:latency} but is not added to the queue, and a background task that meets a full queue is lost. Under the optimal policy, fewer than $10^{-6}$ tasks per episode are placed on a full queue in every system, so this cap has almost no effect on the optimum.

The cloud has no queue, and its latency is set to 110\,ms for the latency metric. To stand for the varying delay of the wide-area network, a cloud task of class $c$ meets its limit with a fixed probability $\pi_c$, which is 0 for the most critical class in every system.

Edge node $i$ finishes one task in a step with probability $d_i$. Other devices add background traffic that the agent does not control. Each background task joins the shortest edge queue. With $b$ background tasks per step on average, the offered edge load is
\begin{equation}
\rho = \frac{1 + b}{\sum_i d_i}.
\label{eq:load}
\end{equation}
All systems are set to $\rho = 1$. The edge then has no spare capacity on average, so queues build up unless some tasks go to the cloud, and the classes compete for the edge. An episode is one run of the system that starts from empty queues and lasts $T=200$ steps.

\begin{table}[t]
\caption{Main notation.}
\label{tab:notation}
\centering
\footnotesize
\setlength{\tabcolsep}{2pt}
\begin{tabular}{@{}ll|ll@{}}
\toprule
$N$, $K$, $T$ & nodes, classes, steps & $p_c$, $L_c$ & share, limit of class $c$ \\
$q_i$, $d_i$ & queue, service prob. & $\pi_c$ & cloud meets $L_c$ \\
$\ell_{0,i}$, $\delta_i$ & time, wait per task & $m_t$ & 1 if task $t$ misses \\
$b$, $\rho$ & background, load & $\beta$, $w_c$, $\kappa$ & bonus, penalty, cloud cost \\
$s_t$, $a_t$ & state, action & $\bar w$, $\alpha$ & flat penalty, scale \\
$\tau_t$ & share of episode left & $e_c$, $v_c$, $J$ & weight, miss rate, cost \\
$\gamma$ & discount factor & $\lambda_c$, $\eta$ & adaptive penalty, step \\
\bottomrule
\end{tabular}
\end{table}

\subsection{Four Systems}
Table~\ref{tab:systems} lists the four systems, and Table~\ref{tab:notation} collects the notation. The nominal system has two equal edge nodes and two classes, critical with a 50\,ms limit and analytics with a 90\,ms limit. An empty queue gives 30\,ms, one waiting task 45\,ms, which still meets the critical limit, and two waiting tasks 60\,ms, which meets only the analytics limit. Background tasks arrive on a fixed schedule, one every five steps (39 in 200 steps, a realized load of 0.996), and the cloud never meets a limit. The three larger systems remove these simplifications: their edge nodes differ in speed, the cloud meets the slower limits with some chance, and the number of background tasks in a step is $\lfloor b \rfloor$ or $\lfloor b \rfloor + 1$, drawn independently with mean $b$. Hetero3-3C and Large4-3C have three classes with limits of 40, 70 and 110\,ms, which stand for control, interactive and analytics tasks.

\begin{table}[t]
\caption{The four systems. All are set to offered edge load $\rho=1$. Latencies in ms. States: number of states of the exact model.}
\label{tab:systems}
\centering
\footnotesize
\setlength{\tabcolsep}{2.5pt}
\begin{tabular}{@{}lcccc@{}}
\toprule
 & Nominal & Hetero3 & Hetero3-3C & Large4-3C \\
\midrule
$N$, $K$ & 2, 2 & 3, 2 & 3, 3 & 4, 3 \\
$\ell_{0,i}$ & 30,30 & 20,30,40 & 20,30,40 & 20,25,30,40 \\
$\delta_i$ & 15,15 & 10,15,20 & 10,15,20 & 10,12,15,20 \\
$d_i$ & .6,.6 & .7,.6,.5 & .7,.6,.5 & .7,.6,.6,.5 \\
$b$ & 0.2 & 0.8 & 0.8 & 1.4 \\
Arrivals & fixed & random & random & random \\
$L_c$ & 50,90 & 50,90 & 40,70,110 & 40,70,110 \\
$p_c$ & .5,.5 & .5,.5 & .3,.3,.4 & .3,.3,.4 \\
$e_c$ & 10,2 & 10,2 & 10,4,1 & 10,4,1 \\
$\pi_c$ & 0,0 & 0,.5 & 0,.2,.6 & 0,.2,.6 \\
States & 32,400 & 291,600 & 437,400 & 3,936,600 \\
\bottomrule
\end{tabular}
\end{table}

\subsection{Decision Process}
\label{sec:steps}
One step runs as follows, numbered as in Fig.~\ref{fig:architecture}. (1) A device sends a task of class $c_t$. (2) The agent reads the state $s_t = (q_1,\dots,q_N, c_t, \tau_t)$. It holds the queue lengths, the class of the task and the share of the episode still to come, $\tau_t = 1 - t/T$. (3) The agent picks an action $a_t$, which is one of the edge nodes or the cloud. (4) The system finds out whether the task meets its limit. For an edge node it uses Eq.~\eqref{eq:latency} and adds the task to the queue. For the cloud it draws the outcome. Then the background tasks join the shortest queue, with ties going to the lowest node index, and each non-empty queue finishes one task with probability $d_i$, independently of the others. (5) The agent gets the reward $R_t$ of Eq.~\eqref{eq:reward}. Then the next task arrives.

\subsection{Rewards}
\label{sec:rewards}
Let $m_t = 1$ if the task of step $t$ misses its limit and $m_t = 0$ otherwise. The reward of a step is
\begin{equation}
R_t = \beta\,(1-m_t) - w_{c_t}\, m_t - \kappa\,\mathbf{1}[a_t = \text{cloud}],
\label{eq:reward}
\end{equation}
where $\mathbf{1}[\cdot]$ is 1 if its condition holds and 0 otherwise. A task that meets its limit earns the bonus $\beta = 1$. A task that misses costs $w_{c_t}$, the penalty of its class. Every cloud placement costs $\kappa=0.1$, so the agent does not use the cloud without reason. The operator who runs the system gives each class a weight $e_c$ that says how costly a miss of that class is. Three choices of penalties are compared. The weighted reward uses these operator weights as penalties, $w_c = e_c$, which gives 10 and 2 in the nominal system. The class-blind (flat) reward uses one penalty for all classes,
\begin{equation}
\bar w = \sum_c p_c\, e_c ,
\label{eq:flat}
\end{equation}
which is 6 in the nominal system and 4.6 in the three-class systems. Both rewards have the same class-average penalty $\sum_c p_c w_c$. The corrected reward follows from the analysis in Section~\ref{sec:rewardproof} and has the same average penalty.

\subsection{Objective}
For each class, the miss rate $v_c$ is the expected number of misses of that class divided by its expected number of tasks. One number is needed to rank policies. The SLA cost per task is
\begin{equation}
J = \sum_c p_c\, e_c\, v_c ,
\label{eq:cost}
\end{equation}
the expected operator weight lost per task. In the nominal system $J = 5\,v_1 + v_2$, so a policy that misses every critical task and no analytics task has $J=5$. The weighted reward uses the same weights, so $J$ could seem to favor it, but Section~\ref{sec:rewardproof} shows that it does not exactly. The paper also reports the miss rate of every class, the full trade-off curve between the classes in the nominal system and, for the DP policies of the rewards, results for weight ratios from 1:1 to 50:1, so that a reader with other weights can judge the policies.

\section{Method}
\label{sec:method}

\subsection{Exact Model}
\begin{figure*}[t]
\centering
\includegraphics[width=\textwidth]{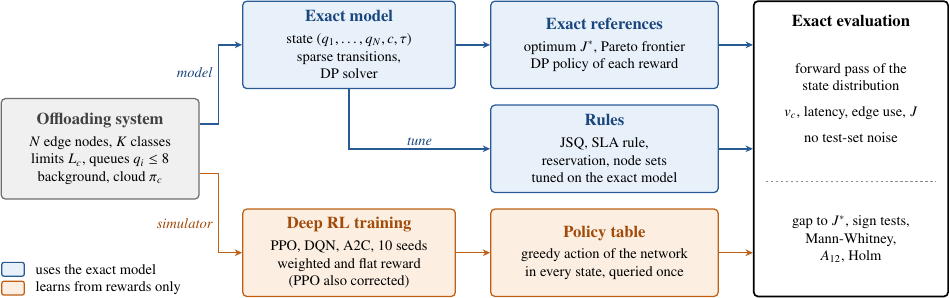}
\caption{Evaluation method. Blue boxes use the exact model, orange boxes learn from rewards only. Each trained network is turned into a table of its greedy action in every state, and the exact evaluation scores this table, the rules and the DP policies with the same forward pass. Only the full rebuilt published design is scored by simulation.}
\label{fig:overview}
\end{figure*}

Fig.~\ref{fig:overview} shows the evaluation method. An exact model is built for each system and used for DP, for tuning the rules and for scoring every policy, while the RL agents learn only from rewards in the simulator.

The state $s_t$ fully describes the system. In the nominal system the background schedule depends only on the step $t$, which $\tau_t$ gives exactly. In the larger systems the background arrivals, like the service events, the cloud outcome and the class of the next task, do not depend on the past. So each system is a finite-horizon Markov decision process, and a deterministic policy that depends only on $s_t$ is optimal among all policies, including those that use the full history \cite[Sec.~4.4]{puterman1994}. There are $9^N$ queue states, $K$ classes and $T=200$ steps, which gives the state counts of Table~\ref{tab:systems}. For each queue state and action, the next queue state follows from the three moves of step (4): the placed task joins its queue, the background tasks join the shortest queue, and each non-empty queue loses one task with probability $d_i$. The transition probabilities $P(s' \mid s,a)$ are stored as one sparse matrix per action.

The open test suite checks the exact model against the simulator. In the nominal system, Hetero3 and Hetero3-3C, exact and simulated miss rates agree within one percentage point over 400 episodes for the optimum, and within 1.5 points for a random policy table. Each extra edge node multiplies the queue states by 9. Table~\ref{tab:scaling} shows the size and cost of the exact model for two to five equal nodes. Five nodes give 23.6 million states, which take 3.4\,s to solve, 6.7\,s for one exact evaluation and 485\,MB of memory. In Large4-3C, solving takes 0.26\,s and one exact evaluation 0.13\,s, while training one PPO agent for 200,704 steps takes 73\,s on the same CPU. For systems of up to four or five nodes the exact benchmark thus costs far less than training a single agent.

\begin{table}[t]
\caption{Size and run time of the exact model for $N$ equal edge nodes ($\ell_{0,i}=30$, $\delta_i=15$, $d_i=0.6$), two classes and $\rho=1$, on an Intel Core Ultra 9 185H. Transitions: stored nonzero transition entries.}
\label{tab:scaling}
\centering
\footnotesize
\setlength{\tabcolsep}{3pt}
\begin{tabular}{@{}rrrrrr@{}}
\toprule
$N$ & States & Transitions & Solve (s) & Evaluate (s) & Memory (MB) \\
\midrule
2 & 32,400 & 1,376 & 0.005 & 0.005 & 57 \\
3 & 291,600 & 32,050 & 0.025 & 0.011 & 59 \\
4 & 2,624,400 & 775,900 & 0.29 & 0.14 & 92 \\
5 & 23,619,600 & 11,302,060 & 3.39 & 6.68 & 485 \\
\bottomrule
\end{tabular}
\end{table}

\subsection{Optimal Policies by Dynamic Programming}
Let $g(s,a)$ be the expected cost of action $a$ in state $s$, and let $V_t(s)$ be the lowest expected cost from step $t$ to the end of the episode. Backward induction computes
\begin{equation}
V_t(s) = \min_{a} \Big[ g(s,a) + \sum_{s'} P(s' \mid s, a)\, V_{t+1}(s') \Big],
\label{eq:bellman}
\end{equation}
with $V_T = 0$. The best action at each step gives the optimal policy \cite[Sec.~4.5]{puterman1994}. Two choices of $g$ are used. With $g$ equal to $e_c$ times the chance of a miss, the total cost is $T$ times $J$, because the expected number of tasks of each class does not depend on the policy, and the result is the optimum. With $g = -\mathbb{E}[R_t]$, the result is the DP policy of the reward. For two classes, Eq.~\eqref{eq:bellman} is also solved with a cost of $\mu/p_1$ for a missed critical task and $(1-\mu)/p_2$ for a missed analytics task, for 101 values of the weighting $\mu$ between 0 and 1. Each solution minimizes $\mu\, v_1 + (1-\mu)\, v_2$ and gives a point on the lower boundary of all reachable pairs of miss rates. Since pairs between two neighboring points can be reached by mixing the two policies, joining the points traces the Pareto frontier, up to the resolution of the 101 weightings.

\subsection{Reward Analysis}
\label{sec:rewardproof}
The reward of Eq.~\eqref{eq:reward} gives $\beta$ for a met limit and $-w_c$ for a miss. So a miss costs $w_c + \beta$ compared with a met limit. This simple fact has large effects.

\emph{Proposition 1.} For any policy, the expected total reward of an episode is
\begin{equation}
\mathbb{E}\Big[\sum_{t} R_t\Big] = T\beta - T\sum_c p_c\,(w_c + \beta)\, v_c - \kappa\, n_\text{cloud},
\label{eq:prop}
\end{equation}
where $n_\text{cloud}$ is the expected number of cloud placements. A policy maximizes the expected reward exactly when it minimizes $\sum_c p_c (w_c + \beta) v_c + \kappa\, n_\text{cloud}/T$.

\emph{Proof.} Each task gives $\beta - (w_{c_t} + \beta)\, m_t - \kappa\,\mathbf{1}[a_t=\text{cloud}]$. Classes are drawn independently of the actions. So the expected number of tasks of class $c$ is $T p_c$ for every policy, and the expected number of misses of class $c$ is $T p_c v_c$. Summing over the steps and taking expectations gives Eq.~\eqref{eq:prop}. The first term does not depend on the policy. \hfill$\blacksquare$

Three consequences follow. First, a learner of the weighted reward ranks misses by $w_c + \beta$, not by $w_c$: in the nominal system the penalties 10 and 2 act as 11 and 3, a ratio of 3.7 instead of the operator's 5. Second, the flat reward ranks all misses equally, and its optimal policy minimizes the overall miss rate plus the cloud term, whatever the operator weights are. Third, the penalties can be corrected. Choosing
\begin{equation}
w_c = \alpha\, e_c - \beta, \qquad \alpha = \frac{\bar w + \beta}{\sum_c p_c e_c},
\label{eq:corrected}
\end{equation}
with the scale factor $\alpha$, makes $w_c + \beta$ proportional to $e_c$ and keeps the average penalty at $\bar w$. This corrected reward ranks policies as $J$ does, apart from the small cloud term. It gives 10.67 and 1.33 in the nominal system, and 11.17, 3.87 and 0.22 in the three-class systems. Setting $\beta = 0$ and $w_c = e_c$ gives the same ranking. Eq.~\eqref{eq:corrected} is useful because some offloading rewards give a positive reward for a task done in time \cite{ullah2023,yamsani2026}. If the weight ratio is large, Eq.~\eqref{eq:corrected} can give a negative penalty. This is still a valid reward, because only $w_c + \beta$ matters. Proposition 1 holds for the undiscounted reward. The learners use a discount factor $\gamma$ below one, which makes a reward $k$ steps ahead count $\gamma^k$ times as much, so Section~\ref{sec:results} also gives the DP policies with discounting.

The proof needs only linearity of expectation, and its value lies in what it exposes. With equal penalties the bonus only rescales the reward, so the shift appears exactly when a met-limit bonus and per-class penalties are used together, the direction that SLA-aware designs take. Among the rewards discussed in Section~\ref{sec:related}, the reward of Yamsani and Chenna Reddy \cite{yamsani2026} has this form: its Eq.~4 adds a success term $\lambda_1 S_i$ for a task finished within its SLA to a penalty $\lambda_2 V_i$, and its Eq.~5 sets only $\lambda_2$ by the Gold, Silver or Bronze tier, although the text mentions larger Gold bonuses. By Proposition 1, a perfect learner of that reward ranks a miss of each tier by $\lambda_2 + \lambda_1$, so the tier ratios that the operator sets are compressed. Ullah et al. \cite{ullah2023} also give a positive reward for a task finished in time. The shaped reward of Mohajer et al. \cite{mohajer2026} has one on-time bonus for all tasks as its only delay term, so by Proposition 1 it acts as a flat penalty, although their objective weights tasks by priority. Other methods use one failure penalty for all tasks or classes \cite{nieto2024,heidari2023,georgiades2026}, rank applications through a graded quality score with fixed priorities \cite{alqerm2021}, or optimize a goal with no deadline term at all \cite{zhang2024,chen2025}. Section~\ref{sec:results} shows that the shift moves the DP policy of the weighted reward up to 10.4\% above the optimum.

\subsection{Baseline Rules}
Four rules serve as baselines. Join-shortest-queue (JSQ) picks the edge node with the fewest waiting tasks and never uses the cloud. The SLA rule knows Eq.~\eqref{eq:latency} and every limit, and picks the edge node with the lowest predicted latency if that latency meets the limit of the task, and the cloud otherwise. The reservation rule adds a latency threshold for each class except the most critical one, and sends a task of such a class to the edge only if the best predicted latency is at most its threshold. This keeps capacity free for critical tasks and mirrors the threshold form of the optimal policy for one pool of equal servers with a common service rate \cite{miller1969}. The node-set rule also lets each class use only a chosen subset of the edge nodes, for example only the slow nodes for analytics. Thresholds and subsets minimize the exact SLA cost, found by exhaustive search for two classes and by coordinate search otherwise. All rules use the true latency model and the exact model, which no learning agent sees.

\subsection{Learning Agents and Tuning}
Three deep RL algorithms are trained: PPO \cite{schulman2017}, A2C, a synchronous variant of A3C \cite{mnih2016}, and DQN \cite{mnih2015}. All come from Stable-Baselines3 2.9.0 \cite{raffin2021} with the Gymnasium 1.3.0 interface \cite{towers2024}. The masked PPO agent of Section~\ref{sec:pubdesign} uses MaskablePPO from sb3-contrib 2.9.0 \cite{raffin2021}. Agents are tested with their greedy action, the action that the network rates best, without random exploration. PPO and A2C update in blocks of collected steps, so a stated budget is reached at the next full block. With blocks of 1,024 or 2,048 steps, 100,000 steps become 100,352 and 200,000 steps become 200,704.

Every algorithm gets the same tuning budget, so no reward or algorithm is helped by better settings. Six configurations are tried for each algorithm: a base setting and five drawn at random from a fixed grid. The base setting is the library default, except that the base DQN updates its target network every 500 steps and explores over 30\% of training. The grid covers the learning rate, the discount factor $\gamma \in \{0.99, 0.999\}$, two hidden layers of 64 or 128 units and three or four further settings of each algorithm. Each configuration is trained with the weighted and the flat reward, with two tuning seeds (100 and 101) that are not used in the main runs, for 50,000 steps rounded up to a full update block. For each reward, the configuration with the highest exact expected total reward of an episode, averaged over the two tuning seeds, is chosen. The SLA cost is never used for this choice, except for the constrained agent of Section~\ref{sec:lagrangian}, and the exact model is used only to rank settings, never during learning. Both rewards chose the same setting for each algorithm, and the corrected reward uses the setting of the weighted reward. The chosen PPO setting has learning rate $10^{-3}$, 1,024 steps per update, batch 128, no entropy bonus, $\gamma = 0.99$ and two hidden layers of 128 units. The chosen DQN and A2C settings are listed with the code.

With the chosen settings, each agent is trained with 10 seeds (0 to 9) for 100,000 steps in the nominal system. The exact model scores every agent every 5,000 steps, which gives learning curves without test noise. In the larger systems PPO is trained with 10 seeds and all three rewards, for 100,000 steps in Hetero3 and Hetero3-3C and 200,000 steps in Large4-3C. The rules are tuned for each system, so PPO is tuned for each system too. The three PPO settings with the highest mean expected return over both rewards in the nominal tuning, which are the chosen setting, the library default and one more, are trained in each larger system with the two tuning seeds and the full budget. For each reward, the one with the highest exact expected return is chosen. The nominal setting was chosen in every larger system, except for the flat reward in Large4-3C, where the default was chosen. To test whether the gap in the larger systems comes from the training budget, weighted and corrected PPO are also trained in Large4-3C for 2,000,000 steps (2,000,896 with full blocks), ten times the main budget, with the same setting and seeds 0 to 9, and the exact model scores them every 50,000 steps.

\subsection{Published Reward Designs}
\label{sec:pubdesign}
The flat reward keeps the key feature of the reward of Nieto et al. \cite{nieto2024}: every failed task gets the same penalty. Their reward has no bonus for a met limit and adds an energy term, which the flat reward does not copy. The design of Yamsani and Chenna Reddy \cite{yamsani2026} is rebuilt in three parts. First, the penalty $\lambda_c$ of class $c$ starts at $e_c$ and is updated after every task of class $c$ as $\lambda_c \leftarrow \max(0, \lambda_c + \eta\,(V_c - V^\ast))$, following Eq.~10 of \cite{yamsani2026}. Here $V_c$ is the miss rate of class $c$ over its last 200 tasks, $V^\ast = 0.05$ is the target and $\eta$ is the step size. The target comes from their global 95\% compliance target. Their paper gives no start value, window length or value of $\eta$, so the start at $e_c$, the floor at zero and the window of 200 tasks are added here, and $\eta = 0.001$ and $\eta = 0.01$ are tested. Second, the state gets two risk features in place of their risk score of execution time against deadline, queue length and past violation rate: the predicted latency of each edge node divided by the limit of the task, and the recent miss rate of each class. Third, their pruning uses risk and load thresholds with no given values. Here, edge nodes whose predicted latency misses the limit are removed by invalid action masking \cite{huang2022}, and the cloud always stays available. A second version keeps only the adaptive penalties. The agent is PPO, or masked PPO when actions are pruned, with the tuned setting of the weighted reward. Their network and their resource pool are not rebuilt.

\subsection{Constrained Baseline}
\label{sec:lagrangian}
The rebuilt design uses targets that no policy can meet (Section~\ref{sec:results}), so a PPO-Lagrangian agent \cite{ray2019,tessler2019} with reachable targets is added. It treats every class except the least critical one as a constraint, $v_c \le \bar v_c$, where $\bar v_c$ is the miss rate of the optimum, so a feasible policy exists by construction. The reward of a step is $\beta(1-m_t) - \lambda_{c_t} m_t - \kappa\,\mathbf{1}[a_t=\text{cloud}]$, where $\lambda_c$ is fixed at $e_c$ for the least critical class and is a Lagrange multiplier for the other classes. After each PPO update block, each multiplier, started at $e_c$, is updated as $\lambda_c \leftarrow \max(0, \lambda_c + \eta\,(\hat v_c - \bar v_c))$ with the miss rate $\hat v_c$ of the last block. The PPO setting is that of the weighted reward. Because the return changes with the multipliers, the step size $\eta \in \{0.1, 1, 10\}$ is chosen on the two tuning seeds by the exact SLA cost. The agent is trained with 10 seeds in the nominal system and in Large4-3C, with the same budgets as the fixed rewards.

\subsection{Exact Evaluation and Statistics}
A trained network is turned into a table by asking it for its greedy action in every state once. The time share $\tau_t$ uses the same 32-bit numbers as the simulator. The exact model then moves the probability of each state forward from empty queues through the 200 steps and adds up the expected misses, latency and edge placements of each class, which gives the exact $v_c$, mean latencies, edge use and $J$. Only the full published design, whose features are not in the state, is scored by simulating 500 held-out episodes. On one core of an Intel Core Ultra 9 185H, solving the DP for Large4-3C takes 0.3\,s and one exact evaluation 0.1\,s, while training one PPO agent for 200,704 steps takes 1.2\,min, so in the systems studied here the optimum is cheaper to compute than one trained agent.

For learning agents the paper reports the mean over seeds with a 95\% percentile bootstrap confidence interval from 10,000 resamples \cite{efron1993}. Two groups of agents are compared with the two-sided Mann-Whitney U test \cite{mann1947}. The effect size is the Vargha-Delaney $A_{12}$ \cite{vargha2000}, the chance that a seed of the first group has a higher value than a seed of the second, with ties counted as one half. The $p$-values of each family of tests are corrected with the Holm method \cite{holm1979}. In the nominal system this family is the 15 reward comparisons of PPO, DQN and A2C on $J$, $v_1$ and $v_2$. An agent is compared with a rule by a two-sided sign test over the 10 seeds.

\subsection{System Changes}
\label{sec:shifts}
Five changes of the nominal system test what happens when the system differs from the one that rules and agents were built for. The wait per queued task rises to 20\,ms, service slows to $d_i = 0.55$, background traffic rises to 0.3 tasks per step or becomes random with the same mean, or the cloud meets the analytics limit with probability 0.5. Rules and agents are first used unchanged. Then the rules are tuned again, and the nominal PPO agents train for another 25,600 steps in the changed system.

\section{Results}
\label{sec:results}

\subsection{Rewards Under a Perfect Learner}
Table~\ref{tab:dprewards} gives the DP policy of each reward in each system. In the nominal system the weighted reward is 1.0\% above the optimum and the corrected reward reaches it. In Hetero3-3C the bonus moves the weighted reward 10.4\% above the optimum, while the corrected reward is within 0.1\% of it. Over all four systems the corrected reward is at most 0.13\% above the optimum. The small remaining gap comes from the cloud term. The flat reward is 65\% to 118\% above the optimum in every system. With the discount factor 0.99 that the PPO and DQN agents use, the corrected reward stays within 0.83\% of the optimum in the first three systems and within 4.1\% in Large4-3C. In the three larger systems the flat reward gets worse with discounting, up to 174\% above the optimum.

\begin{table}[t]
\caption{Exact SLA cost of the optimum, and distance (\%) of the DP policy of each reward to it. W: weighted, C: corrected, F: flat. $\gamma$: discount factor.}
\label{tab:dprewards}
\centering
\footnotesize
\setlength{\tabcolsep}{2.6pt}
\begin{tabular}{@{}lccccccc@{}}
\toprule
 & & \multicolumn{3}{c}{$\gamma = 1$} & \multicolumn{3}{c}{$\gamma = 0.99$} \\
\cmidrule(lr){3-5} \cmidrule(l){6-8}
System & Optimum & W & C & F & W & C & F \\
\midrule
Nominal & 0.2614 & 1.0 & 0.0 & 95.2 & 3.3 & 0.0 & 95.2 \\
Hetero3 & 0.0899 & 1.1 & 0.1 & 64.6 & 1.7 & 0.8 & 115.5 \\
Hetero3-3C & 0.0839 & 10.4 & 0.0 & 116.2 & 10.5 & 0.1 & 169.7 \\
Large4-3C & 0.0777 & 4.9 & 0.1 & 118.2 & 5.5 & 4.1 & 174.0 \\
\bottomrule
\end{tabular}
\end{table}

When the weight ratio $e_1 : e_2$ of the nominal system is changed from 1 to 50, with the average penalty fixed at 6 and $J$ computed with the changed weights, the flat reward is optimal only for equal classes, and its gap grows from 17\% at a ratio of 2 to 95\% at 5 and 589\% at 50. The weighted reward stays within 1.5\% up to a ratio of 10 but reaches 27\% at 50, because the bonus becomes large against the small analytics penalty. The corrected reward is within 0.01\% of the optimum up to a ratio of 20 and 1\% above it at 50.

\subsection{Trained Agents in the Nominal System}
Table~\ref{tab:main} gives every policy in the nominal system after 100,000 training steps. PPO with the weighted reward misses 2.9\% of critical tasks and 14.4\% of analytics tasks, and its SLA cost of 0.291 is 11\% above the optimum. PPO with the flat reward misses 11.3\% of critical tasks and almost no analytics tasks, and its SLA cost of 0.566 is 117\% above the optimum. DQN shows the same pattern, with 0.330 and 0.550. Every PPO and DQN agent sends at least 91\% of tasks to the edge, so none of them avoids misses by hiding tasks in the cloud.

\begin{table*}[t]
\caption{Nominal system after 100,000 training steps. Learning agents: mean over 10 seeds, with a 95\% bootstrap confidence interval for the SLA cost. Rules and DP policies: exact values. All values come from the exact model. $v_1$, $v_2$: miss rate of critical and analytics tasks. All: overall miss rate. $\bar\ell_1$, $\bar\ell_2$: mean latency of critical and analytics tasks. Edge: share of tasks placed on an edge node. Gap: distance to the optimum. Beats: seeds in which the agent has a lower SLA cost than the tuned reservation rule.}
\label{tab:main}
\centering
\footnotesize
\setlength{\tabcolsep}{2pt}
\begin{tabular}{@{}lccccccccc@{}}
\toprule
Policy & $v_1$ (\%) & $v_2$ (\%) & All (\%) & $\bar\ell_1$ (ms) & $\bar\ell_2$ (ms) & Edge (\%) & SLA cost $J$ & Gap & Beats \\
\midrule
JSQ & 51.8 & 14.4 & 33.1 & 61.5 & 61.5 & 100.0 & 2.735 & 946\% & \\
SLA rule & 11.7 & 0.0 & 5.8 & 45.4 & 39.7 & 94.2 & 0.583 & 123\% & \\
Reservation rule & 1.3 & 24.1 & 12.7 & 34.4 & 49.3 & 87.3 & 0.304 & 16\% & \\
\midrule
A2C flat & 13.5 & 0.2 & 6.9 & 46.0 & 42.2 & 94.1 & 0.678 [0.608, 0.763] & 160\% & 0/10 \\
A2C weighted & 18.1 & 3.9 & 11.0 & 45.1 & 46.9 & 93.7 & 0.946 [0.678, 1.247] & 262\% & 0/10 \\
DQN flat & 11.0 & 0.1 & 5.6 & 44.7 & 43.5 & 94.5 & 0.550 [0.540, 0.563] & 110\% & 0/10 \\
DQN weighted & 4.5 & 10.6 & 7.5 & 39.1 & 44.2 & 92.5 & 0.330 [0.309, 0.352] & 26\% & 2/10 \\
PPO flat & 11.3 & 0.2 & 5.8 & 44.9 & 40.8 & 94.3 & 0.566 [0.556, 0.577] & 117\% & 0/10 \\
PPO weighted & 2.9 & 14.4 & 8.7 & 37.0 & 44.6 & 91.4 & 0.291 [0.278, 0.307] & 11\% & 8/10 \\
PPO corrected & 2.7 & 14.8 & 8.8 & 36.9 & 44.7 & 91.3 & 0.284 [0.277, 0.296] & 9\% & 9/10 \\
\midrule
DP flat & 10.2 & 0.0 & 5.1 & 44.3 & 46.6 & 94.9 & 0.510 & 95\% & \\
DP weighted & 2.4 & 14.3 & 8.3 & 37.4 & 46.4 & 91.7 & 0.264 & 1\% & \\
DP corrected & 2.1 & 15.8 & 8.9 & 36.9 & 46.6 & 91.1 & 0.261 & 0\% & \\
Optimum (DP) & 2.1 & 15.8 & 8.9 & 36.9 & 46.6 & 91.1 & 0.261 & 0\% & \\
\bottomrule
\end{tabular}
\end{table*}

For PPO and DQN, every weighted seed has a lower SLA cost and fewer critical misses than every flat seed, and more analytics misses ($A_{12} = 0.00$ for $J$ and $v_1$, $A_{12} = 1.00$ for $v_2$, Holm-corrected $p<0.001$). The corrected reward does not differ significantly from the weighted reward for PPO ($p=1.000$), since its DP policy is only 1.0\% better, far below the spread between seeds. A2C stays far from the optimum, 160\% to 262\% above it and worse on average than the SLA rule. Its two rewards do not differ significantly in SLA cost ($p=1.000$), and only the analytics miss rate differs significantly.

JSQ never uses the cloud and misses 51.8\% of critical tasks. The SLA rule misses 11.7\% of critical tasks, and weighted PPO and weighted DQN beat it in all 10 seeds (sign test, $p=0.002$). The tuned reservation rule sends an analytics task to the edge only when a queue is empty and is 16\% above the optimum. Corrected PPO beats it in 9 of 10 seeds ($p=0.021$) and weighted PPO in 8 of 10 seeds ($p=0.109$), while weighted DQN beats it in only 2 of 10 seeds ($p=0.109$).

\begin{figure}[t]
\centering
\includegraphics[width=\columnwidth]{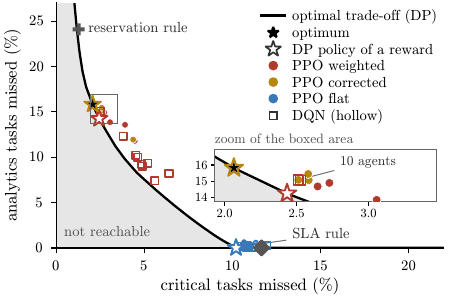}
\caption{Trade-off between critical and analytics misses in the nominal system. The black curve is the optimal trade-off found by DP for 101 weightings, and no policy can reach the gray area below it, up to the resolution of these weightings. Filled circles are PPO agents and hollow squares DQN agents, one per seed, after 100,000 steps. Colors show the reward. The black star is the optimum, and outlined stars are the DP policies of the rewards. The star of the corrected reward surrounds the black star. Gray markers are the SLA rule and the reservation rule. The inset enlarges the boxed area, where ten agents share one point.}
\label{fig:frontier}
\end{figure}

Fig.~\ref{fig:frontier} places every PPO and DQN agent against the optimal trade-off. The weighted and corrected PPO agents lie close to the frontier, near the optimum, while the flat agents lie near its right end, where almost no analytics tasks are missed. They are not bad policies in general: flat PPO and flat DQN have the lowest overall miss rates of all agents, 5.8\% and 5.6\% against 8.7\% for weighted PPO.

\begin{figure*}[t]
\centering
\includegraphics[width=\textwidth]{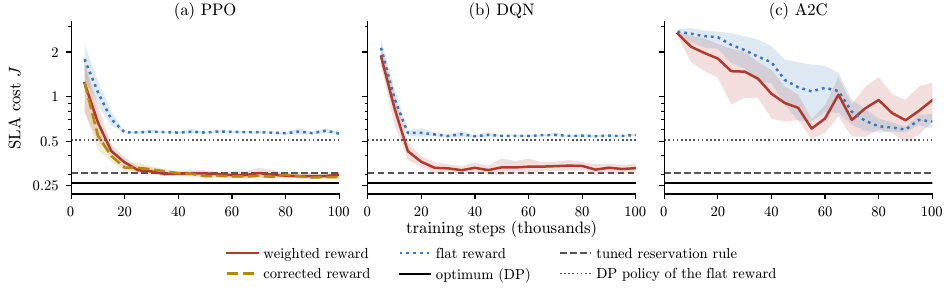}
\caption{Learning curves in the nominal system. Each agent is scored by the exact model every 5,000 steps. Lines are means over 10 seeds and shaded bands 95\% bootstrap confidence intervals. The y-axis is logarithmic. Horizontal lines mark the optimum, the tuned reservation rule and the DP policy of the flat reward.}
\label{fig:curves}
\end{figure*}

On the learning curves in Fig.~\ref{fig:curves}, weighted PPO reaches the level of the tuned reservation rule at 35,000 steps and stays at or below it, within 0.001, and corrected PPO reaches it at 40,000 steps and stays below it. Flat PPO and flat DQN level off 6\% to 15\% above the DP policy of their reward, and A2C is still far from the other two at 100,000 steps.

\subsection{Comparison with Published Designs}
Table~\ref{tab:pub} gives the rebuilt design of Yamsani and Chenna Reddy \cite{yamsani2026}. With risk features, action pruning and adaptive tiers, PPO reaches an SLA cost of 0.360 with $\eta = 0.001$ and 0.383 with $\eta = 0.01$, 38\% and 47\% above the optimum, and 0.414 with adaptive tiers alone. All three versions have a lower SLA cost than the flat reward and a higher one than the fixed weighted reward ($A_{12} \ge 0.94$ against the weighted and $A_{12}=0.00$ against the flat reward, Holm-corrected Mann-Whitney $p < 0.001$ in all six tests). None of them beats the tuned reservation rule in any seed.

\begin{table}[!t]
\caption{Rebuilt design of \cite{yamsani2026} in the nominal system, PPO, 100,000 steps, 10 seeds. full: risk features, pruning and adaptive tiers. tiers only: adaptive tiers alone. $\eta$: step size. Gap: distance to the optimum. $\lambda_1, \lambda_2$: mean penalty of the critical and the analytics tier at the end of training. The full design is scored by 500 simulated held-out episodes (standard error of the SLA cost of each seed at most 0.0103). All other rows are exact. Last two rows: PPO with the fixed rewards.}
\label{tab:pub}
\centering
\footnotesize
\setlength{\tabcolsep}{1.5pt}
\begin{tabular}{@{}lccccc@{}}
\toprule
Variant & $\eta$ & SLA cost $J$ & Gap & $v_1$ (\%) & $\lambda_1, \lambda_2$ \\
\midrule
full & 0.001 & 0.360 [0.339, 0.382] & 38\% & 5.6 & 11.0, 4.9 \\
full & 0.01 & 0.383 [0.357, 0.417] & 47\% & 6.3 & 28.8, 15.8 \\
tiers only & 0.01 & 0.414 [0.393, 0.440] & 58\% & 7.1 & 37.8, 24.1 \\
\midrule
weighted & & 0.291 [0.278, 0.307] & 11\% & 2.9 & 10, 2 \\
flat & & 0.566 [0.556, 0.577] & 117\% & 11.3 & 6, 6 \\
\bottomrule
\end{tabular}
\end{table}

Proposition 1 helps to explain this. The design raises the penalty of a tier whenever its recent miss rate is above 5\%, but in the nominal system no policy meets both 5\% targets. The DP solution for the weighting $\mu = 0.62$ of the frontier shows that every policy has $0.62\, v_1 + 0.38\, v_2 \ge 0.059$, while two miss rates of at most 5\% would give at most 0.05. Both penalties end above their start values, and the analytics penalty grows the most in relative terms, from 2 to between 4.9 and 24.1. With the bonus of 1, the final penalties act as a ratio of about 1.5:1 to 2.0:1 instead of the operator's 5:1, and the critical miss rate of the agents lies between that of the weighted and the flat reward. The convergence proof of the Lagrangian method of \cite{tessler2019} assumes that every local minimum of the constraint cost meets the limit, which needs such a policy to exist.

\subsection{Larger Systems}
\label{sec:larger}
\begin{figure*}[t]
\centering
\includegraphics[width=\textwidth]{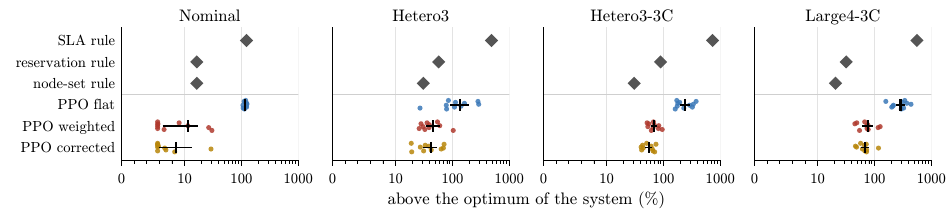}
\caption{Distance to the optimum of every policy in the four systems. The axis is linear below 10\% and logarithmic above. Diamonds are the tuned rules. Dots are PPO seeds, and black marks show the mean (vertical tick) and the 95\% bootstrap confidence interval (horizontal line).}
\label{fig:systems}
\end{figure*}

Table~\ref{tab:scen} and Fig.~\ref{fig:systems} give the results in the larger systems. The tuned reservation rule is much further from the optimum here than in the nominal system, 32\% to 89\% above it, and the node-set rule is 21\% to 31\% above it.

\begin{table}[!t]
\caption{Larger systems. PPO tuned for each system, mean over 10 seeds with 95\% bootstrap interval. Optimum and rules: exact. Above: distance to the optimum, computed from unrounded costs. $v_c$: miss rate of each class in \%. Beats: seeds in which PPO beats the node-set rule.}
\label{tab:scen}
\centering
\footnotesize
\setlength{\tabcolsep}{1.2pt}
\begin{tabular}{@{}lcccc@{}}
\toprule
Policy & SLA cost $J$ & Above & $v_c$ (\%) & Beats \\
\midrule
\multicolumn{5}{@{}l}{\emph{Hetero3}, optimum 0.0899} \\
\, Optimum (DP) & 0.090 & 0\% & 0.4, 6.9 & \\
\, SLA rule & 0.527 & 486\% & 10.5, 0.0 & \\
\, Reservation & 0.141 & 57\% & 1.3, 7.6 & \\
\, Node-set & 0.117 & 31\% & 0.6, 8.9 & \\
\, PPO flat & 0.213 [0.171, 0.262] & 137\% & 3.5, 3.8 & 1/10 \\
\, PPO weighted & 0.130 [0.120, 0.144] & 45\% & 1.2, 7.0 & 3/10 \\
\, PPO corrected & 0.127 [0.117, 0.138] & 41\% & 1.1, 7.4 & 3/10 \\
\multicolumn{5}{@{}l}{\emph{Hetero3-3C}, optimum 0.0839} \\
\, Optimum (DP) & 0.084 & 0\% & 1.2, 0.2, 11.7 & \\
\, SLA rule & 0.697 & 731\% & 23.2, 0.2, 0.0 & \\
\, Reservation & 0.158 & 89\% & 1.5, 0.5, 27.0 & \\
\, Node-set & 0.110 & 31\% & 1.8, 0.6, 12.3 & \\
\, PPO flat & 0.287 [0.250, 0.328] & 243\% & 7.4, 4.4, 3.4 & 0/10 \\
\, PPO weighted & 0.141 [0.134, 0.148] & 68\% & 2.6, 1.7, 10.6 & 0/10 \\
\, PPO corrected & 0.130 [0.124, 0.137] & 55\% & 1.9, 1.8, 13.1 & 0/10 \\
\multicolumn{5}{@{}l}{\emph{Large4-3C}, optimum 0.0777} \\
\, Optimum (DP) & 0.078 & 0\% & 1.0, 0.2, 11.5 & \\
\, SLA rule & 0.511 & 557\% & 17.0, 0.1, 0.0 & \\
\, Reservation & 0.102 & 32\% & 1.0, 0.2, 17.3 & \\
\, Node-set & 0.094 & 21\% & 1.1, 0.5, 13.9 & \\
\, PPO flat & 0.300 [0.262, 0.341] & 286\% & 7.7, 5.0, 2.6 & 0/10 \\
\, PPO weighted & 0.137 [0.125, 0.149] & 76\% & 2.6, 1.7, 9.9 & 0/10 \\
\, PPO corrected & 0.131 [0.123, 0.141] & 68\% & 2.6, 1.0, 10.5 & 0/10 \\
\bottomrule
\end{tabular}
\end{table}

The effect of the reward carries over to every larger system. The weighted reward lowers the SLA cost of PPO against the flat reward from 0.213 to 0.130 in Hetero3, from 0.287 to 0.141 in Hetero3-3C and from 0.300 to 0.137 in Large4-3C. All three differences are significant after the Holm correction over the nine reward comparisons ($p \le 0.012$), and in the three-class systems every weighted seed is below every flat seed. The corrected reward gives 0.127, 0.130 and 0.131 and does not differ significantly from the weighted reward in any system.

The trained agents are still far from the optimum, 41\% to 76\% above it on average for the weighted and corrected rewards, although they beat the SLA rule in every seed. Weighted and corrected PPO beat the tuned reservation rule in 8 and 7 of 10 seeds in Hetero3 ($p=0.109$ and $0.344$) and in 9 and 10 of 10 seeds in Hetero3-3C ($p=0.021$ and $0.002$), but in no seed in Large4-3C. The node-set rule is better than every agent in Hetero3-3C and Large4-3C and than 7 of 10 seeds of each per-class reward in Hetero3. The optimum does not gain its edge mainly by planning around the end of the episode: a version that uses its step-100 action at every step is only 1.8\% to 4.4\% above it in the larger systems.

\emph{Where the gap comes from.} Since $J$ is a weighted sum of the miss rates in Eq.~\eqref{eq:cost}, the gap of an agent to the optimum splits into one part for each class. For weighted and corrected PPO, the critical class gives 46\% to 97\% of the gap in the larger systems, and 81\% and 89\% in Large4-3C. There the agents miss 2.6\% of critical tasks, against 1.0\% for the optimum, while weighted PPO misses fewer tasks of the least valued class than the optimum in both three-class systems. The critical tasks of the agents also take longer, 31.4\,ms and 30.8\,ms on average in Large4-3C against 28.0\,ms for the optimum. The agents thus learn the order of the classes, as Proposition 1 predicts, but protect critical tasks less well than the optimum.

\subsection{Longer Training and a Constrained Baseline}
\label{sec:longrun}
\begin{figure}[t]
\centering
\includegraphics[width=\columnwidth]{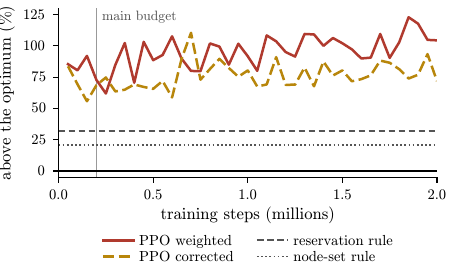}
\caption{Distance to the optimum of PPO in Large4-3C during 2,000,896 training steps, mean over 10 seeds, scored by the exact model every 50,000 steps. The vertical line marks the main budget of 200,704 steps. Gray lines are the tuned reservation rule and the node-set rule.}
\label{fig:longrun}
\end{figure}

After 2,000,896 steps in Large4-3C, weighted PPO reaches an SLA cost of 0.157 [0.146, 0.169] and corrected PPO 0.136 [0.128, 0.143], 102\% and 75\% above the optimum, against 76\% and 68\% after 200,704 steps. Neither change is significant (Holm-corrected Mann-Whitney $p=0.071$ and $0.315$). After 250,000 steps the mean gap of the learning curves moves between 62\% and 123\% for weighted PPO and between 59\% and 110\% for corrected PPO without a downward trend, and no seed beats the node-set rule. The gap is therefore not a matter of training budget (Fig.~\ref{fig:longrun}).

The PPO-Lagrangian agent of Section~\ref{sec:lagrangian}, with the chosen $\eta = 1$, reaches an SLA cost of 0.286 [0.280, 0.293] in the nominal system, 10\% above the optimum, and beats the reservation rule in all 10 seeds ($p=0.002$). Its critical miss rate of 2.6\% stays just above the target of 2.1\%, which only 1 of 10 seeds meets. The multiplier grows only while the miss rate of a block exceeds the target, and only by $\eta$ times that excess, so it lags behind within the training budget: in the second half of training the collected blocks still miss 2.9\% of critical tasks, and $\lambda_1$ rises only from 10 to 13.7. In Large4-3C, with $\eta = 0.1$, it is 68\% above the optimum and meets both targets in no seed. In neither system does it differ significantly from weighted PPO ($p \ge 0.70$). With reachable targets the final multipliers act as a ratio of 4.9:1 after the bonus in the nominal system, close to the operator's 5:1, against 1.5:1 to 2.0:1 for the rebuilt design with unreachable targets.

\subsection{Robustness to System Changes}
Every value here is the distance to the optimum of the changed system. The SLA rule with the nominal latency model is 106\% to 239\% above the optimum. Tuning the reservation rule again picks the same threshold in all five cases, so keeping the nominal threshold costs nothing, and the rule stays 6\% to 16\% above the optimum.

Weighted PPO with its nominal policy is 6\% to 27\% above the new optimum. It beats the unchanged reservation rule in 8 of 10 seeds when the wait per queued task rises and when the background becomes random, where the agent cannot read the background schedule from $\tau_t$, and in no seed under the other three changes. Further training moves the mean of the weighted agents by at most 7 percentage points. Flat PPO stays 99\% to 229\% above the optimum under every change.

The DP policy of the nominal system, used unchanged, is 0.2\% to 23\% above the new optimum. It stays within 3.4\% when tasks or service get slower, but it is 11\% above the optimum when the background becomes random and 23\% above when it rises, where the unchanged reservation rule is 6\% above. An exact policy is thus only as good as the model it is computed from.

\section{Discussion}
\label{sec:discussion}

\emph{Main findings.} The per-class penalty decides which tasks are protected (Proposition 1). In every system, trained PPO agents follow the order that Proposition 1 predicts for perfect learners: the mean SLA cost is lower with the per-class rewards than with the flat reward. Without the optimum, an agent that beats JSQ, an SLA rule and a class-blind version of itself would look strong in all four systems, and its gap of 41\% to 76\% in the larger systems would stay invisible. The rebuilt published design shows that adaptive targets that cannot all be met push the penalties closer.

\emph{Practical guidelines.} If the full model is known and the system is small enough, the DP policy is the best choice, but it should be computed again when the system changes, since a stale DP policy was up to 23\% above the optimum. Tuned rules are simpler, but their quality depends on the system: the tuned reservation rule was 6\% to 16\% above the optimum in the nominal system and its changes and 32\% to 89\% in the larger systems. When a learned agent is needed, use a per-class penalty, fold the bonus for a met limit into the penalties with Eq.~\eqref{eq:corrected}, and check the agent against a strong rule. Reward designs and agents can also be checked first on a reduced model with fewer edge nodes, where the optimum is cheap. With up to four nodes, solving and evaluating take less than a second, against more than a minute to train one agent (Table~\ref{tab:scaling}). Splitting the gap by class, as in Section~\ref{sec:larger}, then shows which class the agent fails to protect.

\emph{Limitations.} The systems are synthetic, with a linear queue model, a cloud described only by the chance to meet each limit and no energy or transmission model, so the results show what the reward does where the answer can be checked exactly, not the size of the effect in a deployed system. Proposition 1 does not depend on these choices, because it holds for any system in which the class of a task does not depend on the actions. The agents use standard networks from one library and a small tuning budget, and Section~\ref{sec:longrun} tests only one longer budget in one system. DQN and A2C were trained only in the nominal system, where PPO was the best learner and A2C stayed far from the optimum, so the larger systems and the system changes use PPO only. The agents see the exact queue lengths, as the DP policy does. Delayed or noisy queue information would make the problem partially observable and would likely widen the gap to the optimum. The exact model is limited to about four or five edge nodes, and for larger systems approximate DP \cite{powell2011adp} could give an approximate reference in its place. For the rebuilt design, start values, window and step size had to be chosen, and its network and resource pool, which its paper does not describe in enough detail, are left out.

\section{Conclusion}
\label{sec:conclusion}
Against an exact optimum, a class-blind penalty stays 65\% to 118\% above the optimum even with a perfect learner, while a per-class penalty corrected for the met-limit bonus comes within 0.13\% of it. Tuned PPO with a per-class penalty comes within 9\% to 11\% of the optimum in the smallest system but stays 41\% to 76\% above it in the larger ones, and ten times more training does not close this gap. Most of this gap comes from critical tasks, so an agent that ranks the classes correctly can still protect the most valuable class too little. Future work will test multi-objective learners, approximate DP for larger systems, partial observation of the queues, and energy and transmission models.

\section*{Code and Data Availability}
All code and results are openly available at \url{https://github.com/tharun-kumar-korinepalli/qos-drl-task-offloading}, together with the policy table of every agent whose policy depends only on the state and the trained PPO networks of the main nominal runs. No other data were used.

\ifCLASSOPTIONcompsoc
\section*{Acknowledgments}
\else
\section*{Acknowledgment}
\fi
The authors used Claude (Anthropic) to edit the text, to check the citations against their sources and to review the code. The authors checked all text, code and results and take full responsibility for the content of this article.

\begin{IEEEbiography}[{\includegraphics[width=1in,height=1.25in,clip,keepaspectratio]{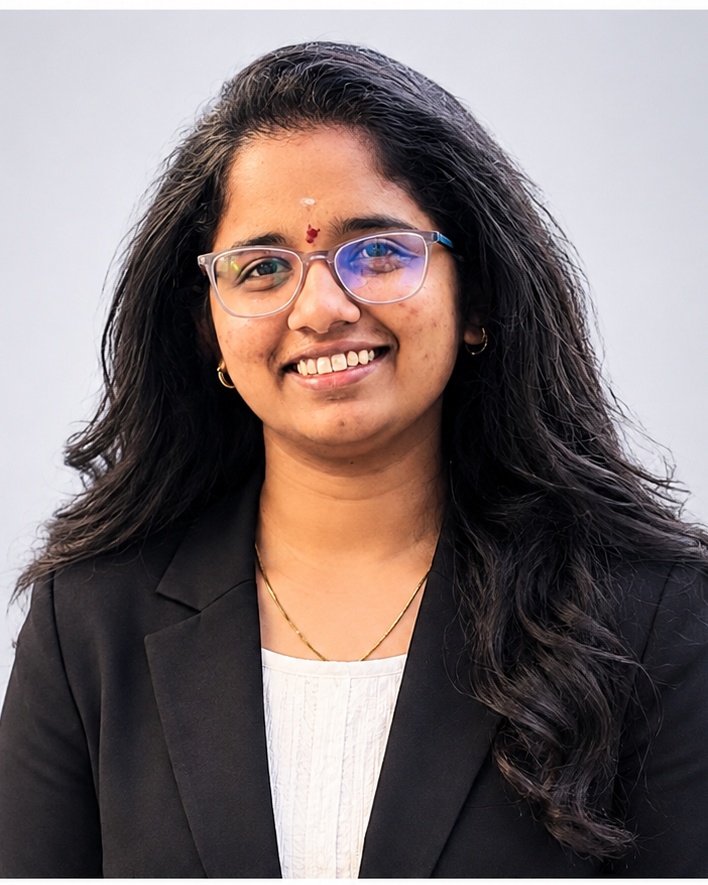}}]{Shilpa Golla}
is currently pursuing the Master of Engineering degree in applied AI for digital production management with the Deggendorf Institute of Technology, Cham, Germany. She worked as an AI Engineer with Deloitte. She is currently a Generative AI Engineer (working student) with Siemens. Her research interests include artificial intelligence, machine learning, reinforcement learning, generative AI, smart production systems and industrial automation, in particular AI methods that improve efficiency, automation and decision-making in manufacturing.
\end{IEEEbiography}

\begin{IEEEbiography}[{\includegraphics[width=1in,height=1.25in,clip,keepaspectratio]{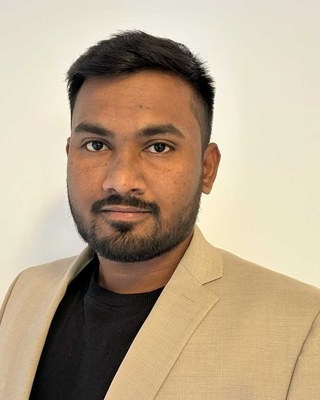}}]{Tharun Kumar Korine Palli}
received the Master of Science degree in artificial intelligence from the Technische Hochschule W\"urzburg-Schweinfurt (THWS), Germany. He worked as a Machine Learning Engineer with Infosys, Hyderabad, India. He is currently a Research Associate with the Technologietransferzentrum Kitzingen (TTZ-KT), THWS, Kitzingen, Germany, where he also supervises master's theses. His research interests include edge computing, quantized models, reinforcement learning, natural language processing and large language models for robot task planning.
\end{IEEEbiography}

\begin{IEEEbiography}[{\includegraphics[width=1in,height=1.25in,clip,keepaspectratio]{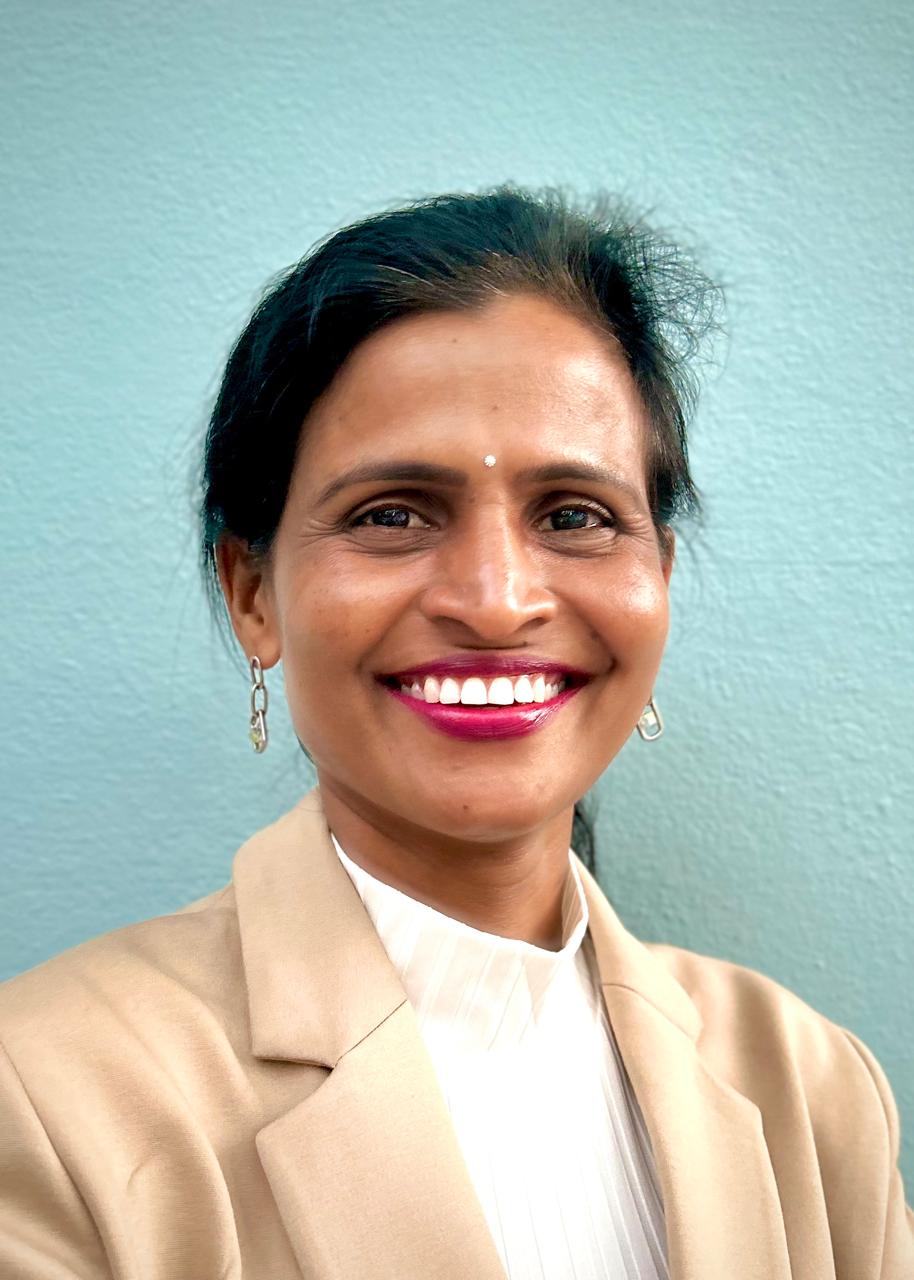}}]{Urmila Bandaru}
received the Master of Technology degree in power electronics from SKD University, India, in 2008, and the Ph.D. degree in electrical engineering from Rayalaseema University, Kurnool, India, in 2017. From 2007 to 2024, she was an Assistant Professor and then an Associate Professor with the Department of Electrical and Electronics Engineering, G. Pulla Reddy Engineering College, Kurnool. She is currently an independent researcher and an academic advisor to SKN Sinhagad College of Engineering, Pandharpur, India. She has authored or coauthored more than 26 indexed publications. Her research interests include multilevel inverters, power electronic converters for electric vehicles, sensorless PMSM control and microgrid optimization.
\end{IEEEbiography}

\end{document}